\documentclass[11pt,a4paper]{article}
\usepackage[hyperref]{acl2021}
\usepackage{times}
\usepackage{latexsym}
\usepackage{amssymb}
\usepackage{amsmath}
\usepackage{booktabs}
\usepackage{graphicx}
\usepackage{algorithm}
\usepackage{algorithmic}
\usepackage{comment}
\usepackage[utf8]{inputenc}
\usepackage{multirow}

\usepackage{cleveref}
\crefname{section}{§}{§§}
\Crefname{section}{§}{§§}

\usepackage{microtype}

\aclfinalcopy 

\title{Target-Oriented Fine-tuning for Zero-Resource \\  Named Entity Recognition}

\author{
  Ying Zhang\textsuperscript{1}\thanks{ \ \ Work was done when Ying Zhang was interning at Pattern Recognition Center, WeChat AI, Tencent Inc, China.}  , 
  Fandong Meng\textsuperscript{2}, 
  \textbf{Yufeng Chen}\textsuperscript{1}\thanks{ \ \ Yufeng Chen is the corresponding author.}, \textbf{Jinan Xu}\textsuperscript{1}
  \ and \textbf{Jie Zhou}\textsuperscript{2}\\
  \textsuperscript{1}Beijing Key Lab of Traffic Data Analysis and Mining, \\Beijing Jiaotong University, Beijing, China \\
  \textsuperscript{2}Pattern Recognition Center, WeChat AI, Tencent Inc, China \\
  \texttt{\{zhying,chenyf,jaxu\}@bjtu.edu.cn} \\
  \texttt{\{fandongmeng,withtomzhou\}@tencent.com} \\
}

\date{}

\begin{document}
\maketitle

\begin{abstract}
Zero-resource named entity recognition (NER) severely suffers from data scarcity in a specific domain or language.
Most studies on zero-resource NER transfer knowledge from various data by fine-tuning on different auxiliary tasks. However, how to properly select training data and fine-tuning tasks is still an open problem.
In this paper, we tackle the problem by transferring knowledge from three aspects, i.e., domain, language and task,  and strengthening connections among them. 
Specifically, we propose four practical guidelines to guide knowledge transfer and task fine-tuning. Based on these guidelines, we design a target-oriented fine-tuning (TOF) framework to exploit various data from three aspects in a unified training manner.
Experimental results on six benchmarks show that our method yields consistent improvements over baselines in both cross-domain and cross-lingual scenarios. Particularly, we achieve new state-of-the-art performance on five benchmarks.
\end{abstract}


\section{Introduction}


Named Entity Recognition (NER) is one of the fundamental tasks in natural language processing.
Recently, zero-resource NER draws more and more attention in recent studies \cite{tackstrom-etal-2012-cross, jia2019cross, bari2020zero, liu2020zeroresource, wu-etal-2020-single}.
This task describes that, in a specific domain or language,  there is no labeled training data for NER. Therefore, zero-resource NER severely suffers from data scarcity.

\begin{figure}
      \centering
      \scalebox{0.45}{
      \includegraphics[width=6.5in]{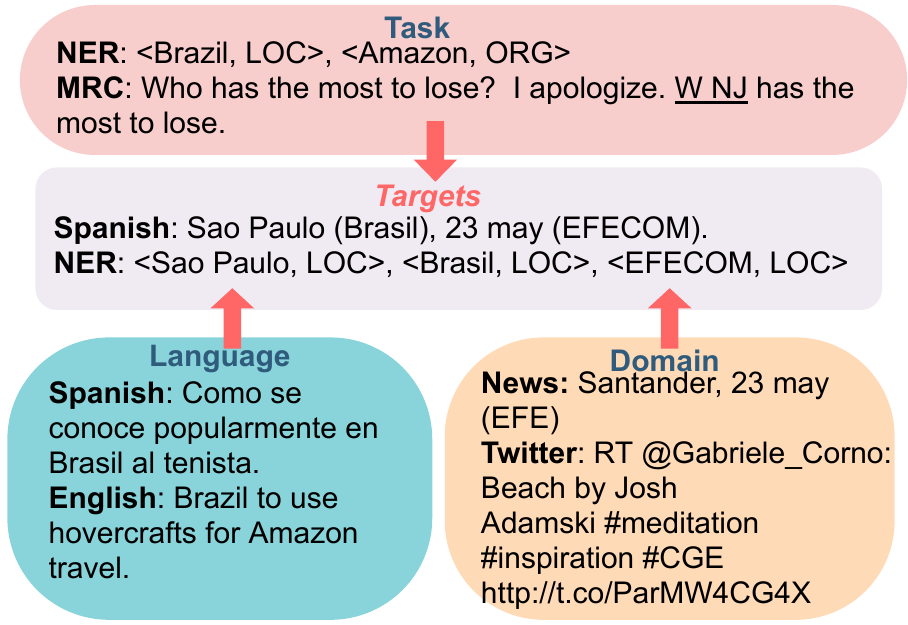}
      }
      \caption{The example of the \textit{Targets} and essential knowledge from three aspects, i.e., Task, Language, and Domain. The middle rectangle denotes an Spanish NER example in news domain, which is referred as the \textit{Targets}. The rounded rectangle above the \textit{Targets} denotes knowledge from different tasks. The bottom left one denotes essential knowledge in Spanish and English languages. The  bottom right one denotes knowledge in News and Twitter domains. } 
      \label{fig:example}
  \end{figure}

As shown in Figure \ref{fig:example}, the ideal training data for zero-resource NER is regarded as the \textit{Targets}, which should satisfy two conditions at the same time: a) in the target domain or language, and b) annotated with NER labels. Thus it is intuitive to augment training data or transfer knowledge from three aspects, i.e., task, language, and domain. The aspect of domain/language can be divided into the source and the target, and the mainstream solution for zero-resource NER is transferring NER annotations from source domains/languages to target ones, e.g., from news to Twitter \cite{strauss2016results} or from English to Spanish \cite{bari2020zero}, where the former is referred as \textit{cross-domain} and the latter as \textit{cross-lingual}.

Based on the mainstream approach, recent researches have conducted further exploration by fine-tuning the contextualized word embeddings on different data. Their results show that only exploiting source labeled data for NER is not enough, due to the discrepancy of domain/language between the source and the target.
To alleviate this problem, some studies focus on utilizing a large amount of target unlabeled data to transfer domain- or language-specific knowledge. AdaptaBERT  \cite{han-eisenstein-2019-unsupervised} fine-tunes the masked language model (MLM) on unlabeled data in the target domain (e.g., social media). Both \citet{pfeiffer-etal-2020-mad} and \citet{vidoni2020orthogonal} add extra components to learn from unlabeled data in the target language (e.g., Spanish). 
Besides, \citet{phang2020english} apply non-NER labeled data in the source language (i.e., English) to transfer knowledge for cross-lingual NER, which suggests that annotations for non-NER tasks (e.g., MRC) are useful for NER task. However, they only exploit non-NER annotations in the source language and ignore that in the target languages (e.g., Spanish). 

Though the aforementioned studies have improved the performance of zero-resource NER in cross-domain or cross-lingual scenarios, there are two main problems in these methods:
a) they conduct knowledge transferring by only considering unlabeled target data and labeled source data, which is insufficient for knowledge transfer. Particularly, they ignore the fact that labeled target data from non-NER tasks is available. 
b) They fine-tune contextualized word embeddings on various auxiliary tasks in a pipeline manner, where each task is performed only once. We argue that the fine-tuning procedure can not capture enough knowledge from various data when trained only once. Besides, it lacks effective strategies to approach the \textit{Targets} closer. 
Target at these issues, we suggest it is necessary to exploit more diverse data and design strategies more oriented to the \textit{Targets}. 

Therefore, we propose four practical guidelines on how to fully exploit available data to alleviate data scarcity.
Concretely, we highlight the necessity of transferring knowledge from three aspects, i.e., task, language, and domain (Guideline-\uppercase\expandafter{\romannumeral1}). 
Then for domain/language, we pay attention to the gap between the source and target data (Guideline-\uppercase\expandafter{\romannumeral2}). For task, we focus on the gap between non-NER tasks and NER (Guideline-\uppercase\expandafter{\romannumeral3}). 
Finally, we emphasize the importance of knowledge fusion between the target domain/language and NER task (Guideline-\uppercase\expandafter{\romannumeral4}).
According to our proposed guidelines, we design a target-oriented fine-tuning (TOF) framework for zero-resource NER to approach the \textit{Targets}. This framework applies three tasks (i.e., MLM, MRC, and NER) to capture the knowledge from above three aspects. It enhances the training with MRC task, pseudo data, and continual learning, respectively.
To validate the effectiveness and superiority of our approaches,  we conduct experiments on six popular benchmarks for zero-resource NER in cross-domain and cross-lingual scenarios. 

Our contributions\footnote{Code and data are publicly available at \url{https://github.com/Yarkona/TOF} } are summarized as follows:
\begin{itemize}
    \item We analyze the key factor of zero-resource NER and propose four practical guidelines to transfer knowledge from three aspects, i.e., Task, Language, and Domain, and strengthen connections among them. 
    \item We design a target-oriented fine-tuning (TOF) framework based on our guidelines to exploit more diverse knowledge and approach the \textit{Targets} closer. 
    \item Experimental results verify the effectiveness of our method in both cross-domain and cross-lingual scenarios on six benchmarks. Particularly, our method achieves the state-of-the-art performance on five benchmarks.
\end{itemize}


\section{Background}
 
 \subsection{Task Definition}
  The goal of zero-resource NER task is to transfer NER knowledge from labeled source data to unlabeled target data. Therefore, we assume that there are three kinds of data available for training: a) NER labeled source data, b) unlabeled target data, and c) non-NER labeled target data (e.g., MRC).

  \subsection{Basic Framework} \label{sec:base_frame}
  Our method is built on AdaptaBERT proposed by \citet{han-eisenstein-2019-unsupervised}. This network is designed for unsupervised domain adaptation on sequence labeling tasks (e.g., NER). 
  A two step fine-tuning approach is applied in AdaptaBERT, and in this section, we will describle it in detail.  

      \paragraph{Step-1: Domain Tuning.} They fine-tune contextualized word embeddings by training a masked language model (MLM) to reconstruct randomly masked tokens. And this is performed on a dataset containing all available target domain data and an equal amount of unlabeled source domain data. 

      \paragraph{Step-2: Task Tuning.} They fine-tune contextualized word embeddings continually and learn the prediction model for the sequence labeling task.
      Following \cite{devlin2018bert}, they build a strong NER system by simply feeding the contextualized embeddings into a linear classification layer. The log probability can be computed by the log softmax,
       \begin{equation} 
            \log p(y_t|\textbf{w}_{1:T}) = \beta_{y_t} \cdot \textbf{x}_t - \log \sum_{y \in \mathcal{Y}} \exp(\beta_y \cdot \textbf{x}_t), 
       \end{equation}
      where contextualized word embedding $\textbf{x}_t$ captures information from the entire sequence $\textbf{w}_{1:T} = {w_1, w_2, ..., w_T}$, and $\beta_y$ is a vector of weights for each tag $y \in Y =\{PER, ORG, LOC, MISC\}$.
      They train the model on labeled source domain data by minimizing the negative conditional log-likelihood of labeled data.

\section{Our Approach}
  For zero-resource NER, we firstly analyze the problem of data scarcity. Then we propose four practical guidelines to guide knowledge transfer from different data, which is adapted to both cross-domain and cross-lingual scenarios. According to these guidelines, we design a target-oriented fine-tuning (TOF) framework for zero-resource NER. 

  \subsection{Problem Analysis}
  The nature of zero-resource NER task is to perform NER with no labeled target domain/language data. And to deal with this task, it is intuitive to transfer essential knowledge from other available data.
  Concretely, when the data satisfies the two conditions at the same time: a) in the target domain/language (e.g., Twitter/Spanish), and b) annotated for the target task (i.e., NER),  we consider it as our \textit{Targets}. While the \textit{Targets} is unavailable under the zero-resource setting, there is abundant data meeting either condition. 
  Therefore, we transfer knowledge from three aspects, i.e., Domain, Language, and Task, as shown in Figure \ref{fig:example}.

  \noindent \textbf{Domain.} It contains knowledge from specific domains (e.g., Twitter). As shown in the bottom right rectangle of Figure \ref{fig:example}, `@Garbriele\_Corno:' is the special expression that only exists in tweets and `\#' is used to highlight something.
  
  \noindent \textbf{Language.} It refers to linguistic knowledge in various languages. For example, the word order of `Como se conoce popularmente en Brasil al tenista' in Spanish is different from its English expression `As the tennis player is popularly known in Brazil'. Besides, the expressions of `Brazil' and `tenista' in English vary from those in Spanish.
  
  \noindent \textbf{Task.} It describes hand-crafted annotations for different tasks, which is expensive and difficult to obtain (e.g., NER and MRC). For example, NER labels \textit{LOC} and \textit{ORG} denote names of locations and organizations, respectively. For MRC task in Figure \ref{fig:example}, `W NJ' is annotated as the answer to question `Who has the most to lose?' .

  Furthermore, we divide domain/language aspect into the source and target. Particularly, NER is regarded as the target task for zero-resource NER.

  \subsection{Four Practical Guidelines}
    Based on our analysis, we propose four practical guidelines on how to fully exploit available knowledge to alleviate data scarcity.

        \textit{Guideline-\uppercase\expandafter{\romannumeral1}:} It is necessary to exploit available knowledge from domain, language, and task.
        

        \textit{Guideline-\uppercase\expandafter{\romannumeral2}:} Bridge the gap between source domains/languages and target domains/languages.     
        

        \textit{Guideline-\uppercase\expandafter{\romannumeral3}:} Bridge the gap between  annotations for non-NER tasks and NER task.
        

        \textit{Guideline-\uppercase\expandafter{\romannumeral4}:} Fuse the knowledge of both the target domain/language and NER task. 
        

     \begin{figure*}
      \centering
      \scalebox{0.78}{
      \includegraphics[width=6.5in]{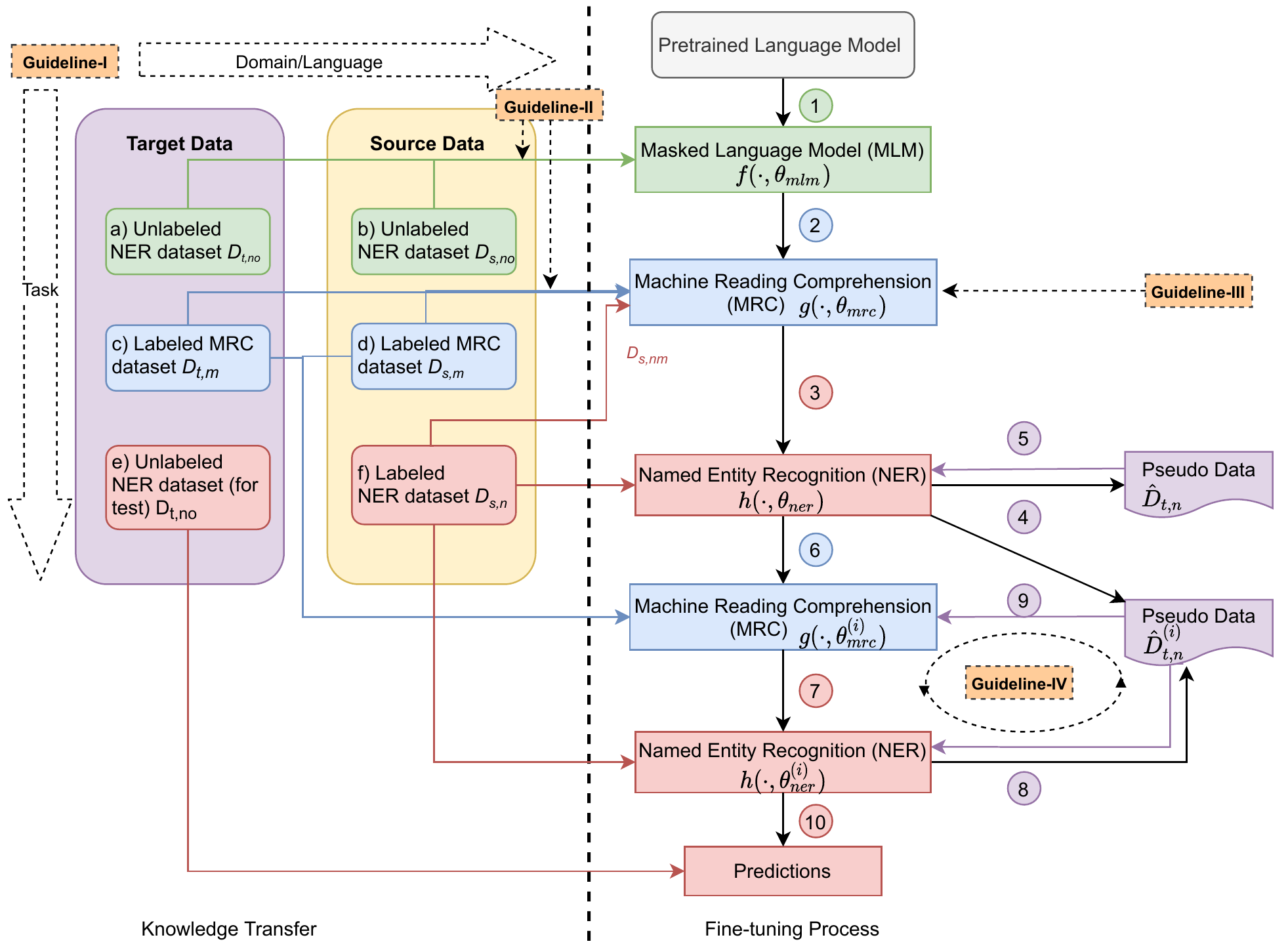}
      }
      \caption{The overall architecture of our Target-Oriented Fine-tuning (TOF) framework. Orange rectangles denote our proposed four guidelines, and dotted lines with arrows denote how they guide \textit{Knowledge Transfer} and \textit{Fine-tuning Process}. Rectangles and solid lines in green, blue, and red color correspond to three fine-tuning tasks (i.e., MLM, MRC, NER), respectively. And the black solid lines with arrows denote the training steps.`Target Data' and `Source Data' denote data in target and source domains/languages, respectively. $i$ in the circle denotes \textit{Step-$i$}.} 
      \label{fig:framework}
  \end{figure*} 
 
  \subsection{Target-Oriented Fine-tuning Framework}

   As shown in Figure \ref{fig:framework}, we design a Target-Oriented Fine-tuning (TOF) framework for zero-resource NER. It contains two components: a) \textit{Knowledge Transfer}, which displays how to transfer not only domain/language but also task knowledge from various data, and b) \textit{Fine-tuning Process}, which demonstrates a flow diagram of the complete fine-tuning process.
   Both components are designed based on our proposed guidelines, and their relations are illustrated in Figure \ref{fig:framework}.
   
   \subsubsection{Knowledge Transfer}
   
   As  the right part of Figure \ref{fig:framework} shows, to transfer both domain/language and task knowledge for the \textit{Targets}, we consider six kinds of corpora: a) unlabeled NER dataset $D_{t,no}$, b) unlabeled NER dataset $D_{s,no}$, c) labeled MRC dataset $D_{t,m}$, d) labeled MRC dataset $D_{s,m}$ , e)unlabeled NER dataset $D_{t,no}$, and f) labeled NER dataset $D_{s,n}$, where \{a), c), e)\} is in the target domain and \{b), d), f)\} is in the source domain. Note that e) is the \textit{Targets} without considering labels. 
              
   According to \textit{Guideline-\uppercase\expandafter{\romannumeral1}}, since there is no available data that satisfies the \textit{Targets}, it is necessary to transfer knowledge relevant to the \textit{Targets} from other data as much as possible. Apart from source NER labeled data,  we not only exploit unlabeled target data, but also utilize non-NER labeled target data. Therefore, three kinds of data are considered as shown in Figure \ref{fig:framework}: for `Target Data', a) unlabeled NER dataset $D_{t,no}$ and c) labeled MRC dataset $D_{t,m}$ ; for `Source Data', and f) labeled NER dataset $D_{s,n}$.      
   
   According to \textit{Guideline-\uppercase\expandafter{\romannumeral2}}, there is discrepancy between the source and target domain/language. To deal with the gap, it is essential to apply fine-tuning tasks on the mixture of the source and target data. Besides, an effective way to bridge the gap is transforming source data into the target format, e.g., translate the source language data into target language. Therefore, we collect b) unlabeled NER dataset $D_{s,no}$ and d) labeled MRC dataset $D_{s,m}$ in the `Source Data'.
  
     \subsubsection{Fine-tuning Process} 
      Based on AdaptaBERT, we novelly introduce a MRC task between \textit{domain-tuning} and \textit{task-tuning} process. Thus, our fine-tuning process contains three fine-tuning tasks as follows.

        \paragraph{Masked Language Model (MLM).} To adapt contextualized word embeddings to both the source and target data, we use MLM \cite{devlin2018bert}. 
        Based on \textit{Guideline-\uppercase\expandafter{\romannumeral2}}, We train the model on a mixture of dataset $D_{t,no}$ and $D_{s,no}$. 
        We use the same strategy with \cite{han-eisenstein-2019-unsupervised} to generate 10 random maskings for each instance.
       
        \paragraph{Machine Reading Comprehension (MRC).}
        Based on the \textit{Guideline-\uppercase\expandafter{\romannumeral3}}, we add a span extraction MRC task, which has three advantages:  a) MRC can enhance the ability of NER model on span extraction and help NER better capture semantic information of different entity types;  b) MRC framework can be used to solve NER task \cite{li-etal-2020-unified} and it becomes a bridge between NER and other tasks; and c) Recent work on framing other tasks as MRC \cite{wu2019coreference,liu2020event}  provides an idea for transferring knowledge from different tasks with a unified framework.
        The MRC model is implemented by feeding the contextualized word embedding of each token $x_t$ into two linear classification layers, respectively. The probability of each token being the start or the end index of a span can be computed as follows:
       \begin{equation}
\begin{split}
            \setlength{\abovedisplayskip}{3pt}
            p_t^{start} =& softmax(W_{start} \cdot \textbf{x}_t), \\
            p_t^{end} =& softmax(W_{end} \cdot \textbf{x}_t), 
            \setlength{\belowdisplayskip}{3pt}
\end{split}
\end{equation}
        where $W_{start}$ and $W_{end}$
        $\in \mathbb{R}$ 
        $^{d_1 \times 2}$ 
        are learnable parameters, and $d_1$ denotes the dimensions of contextualized word embedding. 
        Finally, the model is trained by optimizing the Cross-Entropy loss over $p_t^{start}$ and $p_t^{end}$. 
        According to \textit{Guideline-\uppercase\expandafter{\romannumeral2}}, we train the MRC model on all available MRC data $D_{t,m}$, $D_{s,m}$ and NER data $D_{s,n}$ that is transformed into MRC format $D_{s,nm}$ following \cite{li-etal-2020-unified}.

       \paragraph{Named Entity Recognition (NER).} To fine-tune contextualized word embeddings continually and learn the prediction model, we feed contextualized word embeddings into a linear classification layer and maximize the probability of each token with the ground-truth entity label. Concretely, given an input token sequence $\textbf{x} = \{x_i\}_{i=1}^{N}$ with $N$ words, we firstly feed it into the feature encoder $f_\theta$ to obtain contextualized word embeddings $\textbf{h}=\{h_i\}_{i=1}^N$ for all tokens:
       \begin{equation}
       \setlength{\abovedisplayskip}{3pt}
           \textbf{h}=f_\theta(\textbf{x}),
           \setlength{\belowdisplayskip}{3pt}
       \end{equation}
       where $h_i$ is the feature vector corresponding to the $i$-th token $x_i$ and $f_\theta$ is based on pre-trained language model, i.e., BERT \cite{devlin2018bert}, where $\theta$ denotes model parameters. Then $h_i$ is fed into a linear classification layer with the \textit{softmax} function to predict the probability distribution of entity labels, which is formulated as follows:
       \begin{equation}
       \setlength{\abovedisplayskip}{3pt}
           p(\hat{y}|x_i) = softmax(Wh_i + b), 
           \setlength{\belowdisplayskip}{3pt}
       \end{equation}
       where $\hat{y} \in Y $ with $Y$ being one-hot vectors corresponding to different entity labels, and $\{W, b\}$ denotes learnable parameters. The loss function is defined as the cross entropy between the predicted probability distribution of each entity label and the ground-truth one for each word. We train NER model on $D_{s,n}$ and predict labels on $D_{t,no}$.
      
    \subsubsection{Training}
     A novel training process is proposed to narrow the gap between the knowledge from available data and the \textit{Targets}, which contains three processes, i.e., MRC enhancing, pseudo data enhancing, and continual learning enhancing.
     
     \noindent  \textbf{MRC Enhancing.} We fine-tune contextualized word embeddings by sequentially training the MLM $f(\cdot,\theta_{mlm})$, MRC $g(\cdot,\theta_{mrc})$, and NER $h(\cdot,\theta_{ner})$ at \textit{Step-1$\sim$3} in Figure \ref{fig:framework}.  
       
     \noindent \textbf{Pseudo Data Enhancing.} According to \textit{Guideline-\uppercase\expandafter{\romannumeral4}},  we use the trained NER model (\textit{Step-3}) to generate pseudo labels on NER unlabeled target data $\hat{D}_{t,n}$ (\textit{Step-4}) and then fine-tune NER model $h(\cdot,\theta_{ner}^{(0)})$ continually on generated pseudo-labeled target data at \textit{Step-5}.
  
  \begin{algorithm}[htb]
       \caption{The training process of TOF.}
       \label{alg:framework}
       \begin{algorithmic}[1]
        \REQUIRE
        Dataset $D_{t,no}$, $D_{s,no}$, $D_{t,m}$, $D_{s,m}$, $D_{s,n}$, and $D_{s,nm}$;  
        MLM $f(\cdot;\theta_{mlm})$; 
        MRC $g(\cdot;\theta_{mrc})$; 
        NER $h(\cdot;\theta_{ner})$;
        pre-trained BERT $\theta^{(0)}$;
        Number of pseudo-data  iterations $T$.\\
        \ENSURE $h(\cdot,\theta_{ner}^{(T)})$. \\
        \STATE Initialize $\theta_{mlm}$ = $\theta^{(0)}$
        \STATE Fine-tune $f(\cdot,\theta_{mlm})$ on $\{D_{t,no}, D_{s,no}\}$
        \label{code:fram:step1}
        \STATE  Initialize $\theta_{mrc}$ = $\theta_{mlm}$
        \STATE Fine-tune $g(\cdot,\theta_{mrc})$ on$\{D_{t,m}, D_{s,m},D_{s,nm}\}$
        \label{code:fram:step2}
        \STATE Initialize  $\theta_{ner}$ = $\theta_{mrc}$
        \STATE Fine-tune $h(\cdot,\theta_{ner})$ on  $\{D_{s,n}\}$
        \label{code:fram:step3}
        \STATE Gen pseudo-NER $\hat{D}_{t,n} \leftarrow h(\cdot,\theta_{ner})$on $D_{t,no}$
        \label{code:fram:step4}
         \STATE Initialize $\theta_{ner}^{(0)}$ = $\theta_{ner}$
        \STATE Fine-tune $h(\cdot,\theta_{ner}^{(0)})$ on $\{\hat{D}_{t,n}\}$
        \STATE Gen pseudo-NER $D_{t,n}^{(0)} \leftarrow  h(\cdot,\theta_{ner}^{(0)})$ on $\hat{D}_{t,no}$
        \STATE Gen pseudo-MRC $\hat{D}_{t,m}^{(0)} \leftarrow \hat{D}_{t,n}^{(0)}$ 
        \label{code:fram:step5}
        \FOR{$i = 1 \to T$}
            \STATE{Initialize $\theta_{mrc}^{(i)}$ = $\theta_{ner}^{(i-1)}$}
            \STATE{ Fine-tune  $g(\cdot,\theta_{mrc}^{(i)})$ on $\{D_{t,m}, \hat{D}_{t,m}^{(i-1)} \}$}
            \label{code:fram:step2-i}
            \STATE Initialize $\theta_{ner}^{(i)}$ = $\theta_{mrc}^{(i)}$
            \STATE Fine-tune $h(\cdot,\theta_{ner}^{(i)})$ on $\{\hat{D}_{t,n}^{(i-1)}\}$
           \label{code:fram:step3-i}
           \STATE Gen pseudo-NER $\hat{D}_{t,n}^{(i)} \leftarrow h(\cdot,\theta_{ner}^{(i)})$ on  $D_{t,no}$
           \label{code:fram:step4-i}
           \STATE Gen pseudo-MRC $\hat{D}_{t,m}^{(i)} \leftarrow \hat{D}_{t,n}^{(i)}$
           \label{code:fram:step6-i}
        \ENDFOR
        \label{code:fram:LOOP}
        \STATE Predict $h(\cdot,\theta_{ner}^{(T)})$ on $D_{t,no}$
        \label{code:fram:step7}
        \RETURN $h(\cdot,\theta_{ner}^{(T)})$
       \end{algorithmic}
\end{algorithm}

    \noindent \textbf{Continual Learning Enhancing.} We design a continual learning strategy to make full use of pseudo data and imitate the training procedure on the \textit{Targets}. We continually perform fine-tuning between MRC and NER with considering pseudo data (\textit{Step-6$\sim$7} ), based on the following three considerations: 1) pseudo-labeled target NER data can be refined by the fine-tuned NER model after each iterations, 2) pseudo data is transformed into MRC format, which directly introduces entity type knowledge in target data through definition of MRC questions, and 3) pseudo data participates in both MRC and NER training, which can enhance knowledge connections between two tasks. At \textit{Step-8$\sim$9}, we refine pseudo data with newly fine-tuned NER model and take it as training data.
    After $T$ times iteration, we conduct predictions on unlabeled target data with NER model $h(\cdot,\theta_{ner}^{(T)})$ (\textit{Step-10}).
       
    The training procedure is summarized as \ref{alg:framework}. 
$D_{t,no}$ and $D_{s,no}$ demote unlabeled NER data in the target and source domain/language, respectively.
$D_{t,m}$ and $D_{s,m}$ denote labeled MRC data in the target and source domain/language, respectively.
$D_{t,n}$ and $D_{s,n}$ denote labeled NER data in the target and source domain/language, respectively.
Particularly, $D_{t,n}$ is the pseudo-labeled  NER data in the target domain/language generated by NER model. And we transformed it into the MRC format, as $D_{s,nm}$.
$f(\cdot;\theta_{mlm})$, $g(\cdot;\theta_{mrc})$ and $h(\cdot;\theta_{ner})$ denote the model of MLM, MRC, and NER, respectively. 
Note that `Gen' in Algorithm \ref{alg:framework} denotes the generalize operation. 
    
\section{Experiments}
     \subsection{Data Preparation}
     We take CoNLL03 for English (en) in the news domain as the source dataset for both cross-lingual and cross-domain tasks.
     
     \noindent \textbf{Cross-Lingual}. We consider three NER datasets in target languages: CoNLL03 for German (\textit{de}) \cite{tjong2003introduction}, CoNLL02 for Dutch (\textit{nl}) and Spanish (\textit{es}) \cite{tjong-kim-sang-2002-introduction}. All datasets are labeled with 4 entity types: PER, ORG, LOC, MISC. Each of them is split into training, validation and test sets following \cite{wu2020unitrans}.
     We use three MRC datasets in target languages: MLQA (\textit{es}) \cite{DBLP:journals/corr/abs-1910-07475}, XQuAD (\textit{de}) \cite{Artetxe:etal:2019}, and SQuAD (\textit{en}) \cite{DBLP:journals/corr/RajpurkarZLL16}.
     
     \noindent \textbf{Cross-Domain}. We use three English datasets in target domains: CBS SciTech News dataset \cite{jia2019cross}, short as CBS, in the science and technology news domain, Twitter NER \cite{zhang2018adaptive} and WNUT16 \cite{strauss2016results} in the social media domain. We use two English MRC datasets from news and twitter domains respectively: NewsQA \cite{DBLP:journals/corr/TrischlerWYHSBS16} and TweetQA \cite{DBLP:journals/corr/abs-1907-06292}.
     The statistics of datasets are shown in \ref{tab:datasets} in Appendix \ref{sec:statistics}. 
     
     \begin{table*}[]
      \centering
      \scalebox{0.78}{
      \begin{tabular}{l|cccc|cccc}
      \toprule
        \multirow{2}*{\textbf{Systems}} & \multicolumn{4}{c}{\textbf{cross-lingual}} & \multicolumn{4}{c}{\textbf{cross-domain}} \\
       \cline{2-5} \cline{6-9}
       ~& es & nl & de & avg  & CBS & Twitter & WNUT16 & avg\\
      \midrule
      BERT-ML\cite{DBLP:journals/corr/abs-1912-01389}$^{\dagger}$
      & 76.53 & \textbf{83.35} & 72.44 & 77.44 & - & - & - & -\\
      TSL \cite{wu-etal-2020-single}$^{\dagger}$ 
      & 78.00 & 81.33 & 75.33 & 78.22 & - & - & - & -\\
      UniTrans\cite{wu2020unitrans}$^{\dagger}$ 
      & 79.31 & 82.90 & 74.82 & 79.01  & - & - & - & -\\
      ~ + ensemble & 79.29 & 83.07 & 75.55 & 79.29 & - & - & - & - \\
      mCell LSTM\cite{jia-zhang-2020-multi}$^{\dagger}$ 
      & - & - & - & - & 75.19 & - & - & - \\
      COFEE-MRC\cite{xue2020coarse}$^{\dagger}$
      & - & - & - & - & - & 54.56 & - & -\\
      \midrule
      \multirow{2}*{AdaptaBERT\cite{han-eisenstein-2019-unsupervised}}$^{*}$
      & 75.30 & 78.52 & 70.90 & \multirow{2}*{75.20} & 75.30 & 65.61 & 63.03$^{\ddagger}$ & \multirow{2}*{67.98}  \\
      ~&($\pm$ 0.30) &($\pm$ 0.25)&($\pm$ 0.54)&~&($\pm$ 0.37) &($\pm$ 0.46)&($\pm$ 0.23) &~\\
      \multirow{2}*{~ + translation} & 76.18 & 80.30 & 72.47& \multirow{2}*{76.32} & \multirow{2}*{-} & \multirow{2}*{-} & \multirow{2}*{-} & \multirow{2}*{-}\\
      ~&($\pm$ 0.13) &($\pm$ 0.52)&($\pm$ 0.75)&~&~&~&~&~\\
      \midrule
      \multirow{2}*{TOF(\textit{ours})} & \textbf{80.35} & 82.79 & \textbf{76.57} & \multirow{2}*{\textbf{79.90}} & \textbf{76.41} & \textbf{67.94} & \textbf{67.86} &   \multirow{2}*{\textbf{70.74}} \\
     ~& ($\pm$ 0.29) & ($\pm$ 0.17) & ($\pm$ 0.16) &~&($\pm$ 0.5) & ($\pm$ 0.09) & ($\pm$ 0.27)&~ \\
      \multirow{2}*{~ w/o continual learning} & 79.44 & 81.64 & 76.39 & \multirow{2}*{79.16} & 75.95 & 67.13& 67.70 & \multirow{2}*{70.26} \\
      ~& ($\pm$ 0.08) & ($\pm$ 0.17) &{$\pm$ 0.17} & ~ &($\pm$ 0.38) &($\pm$ 0.05) &($\pm$ 0.08)&~ \\
      \multirow{2}*{~ w/o pseudo data \& w/o continual learning} & 78.32 & 80.56 & 73.61 & \multirow{2}*{77.50} & 75.34 & 66.18 & 66.45 & \multirow{2}*{69.32}\\
      ~& ($\pm$ 1.11) & ($\pm$ 0.35) &{$\pm$ 0.61} & ~ &($\pm$ 0.51) &($\pm$ 0.17) &($\pm$ 0.39)&~ \\
      
      \bottomrule
      \end{tabular}}
      \caption{ 
      Results of our method and previous state-of-the-art methods for zero-resource NER in cross-lingual and cross-domain. 
      `avg' denotes the average of F1 scores (\%) on three benchmarks.
      `$^{\dagger}$' denotes original results reported in their original papers. 
      `$^{*}$' denotes results re-implemented by us. 
      Note that `$^{\ddagger}$' denotes our re-implemented result on WNUT16 and the previous state-of-the-art result on it is 62.8 reported by \citet{han-eisenstein-2019-unsupervised}. }
      \label{tab:overall}
  \end{table*}
  
     \subsection{Data Preprocessing}
      NER datasets are processed in the `BIO' scheme with four entity types, i.e., PER, LOC, ORG, and MISC except for WNUT16. We perform entity span detection task on WNUT16. Since there are ten entity types annotated in WNUT16, it is different from annotations in source domain/language. 
      For MRC datasets, we transform all of them into a unified format following \cite{li-etal-2020-unified} for MRC training. 
      Besides, following \cite{li-etal-2020-unified}, we map the labeled NER datasets to labeled MRC dataset. Concretely, we use the description of each entity for annotators as the query, and each sentence as context. The corresponding answers for each query are entity spans with the same entity type in the sentence. 
     We delete all entity labels on the target data and only use the unlabeled data. We use training and validation sets from the source for training and evaluation, and do predictions on test sets from different target domains/languages.

     \subsection{Implementation Details}
      We use BERT-base and multilingual BERT \cite{devlin2018bert} to initialize contextualized word embeddings in cross-domain and cross-lingual scenarios, respectively. 
      We empirically follow the hyperparameter settings of \cite{han-eisenstein-2019-unsupervised} and \cite{li-etal-2020-unified} except for the learning rate and batch size. Due to the discrepancy between various datasets, we choose the learning rate for Adam \cite{DBLP:journals/corr/abs-1810-12281} optimizer according to the best performance of checkpoints on the validation set. And the batch size is set to 32, 16 and 64 for MLM, MRC and NER, respectively. More hyperparameters for training procedure are listed in Appendix  \ref{sec:hyperparameter}.

     \subsection{Systems}
     We evaluate following systems by entity-level F1 scores \cite{sang2003introduction}. Moreover, we conduct each experiment 5 times and report the mean F1-score.
     
         \noindent \textbf{BERT-ML}. \citet{DBLP:journals/corr/abs-1912-01389} apply the multilingual BERT to cross-lingual NER.
         
         \noindent \textbf{TSL}. \citet{wu-etal-2020-single} propose a teacher-student learning method for cross-lingual NER.
         
         \noindent \textbf{UniTrans}. \citet{wu2020unitrans} unify data transfer and model transfer for cross-lingual NER.
         
         \noindent \textbf{mCell LSTM}. \citet{jia-zhang-2020-multi} design a multi-cell compositional LSTM for cross-domain NER.
         
         \noindent \textbf{COFEE-MRC}. \citet{xue2020coarse} inject coarse-to-fine automatically mined entity knowledge in a pre-trained language model for cross-domain NER.   
         
          \noindent \textbf{AdaptaBERT}. \citet{han-eisenstein-2019-unsupervised}  perform \textit{domain-tuning} and \textit{task-tuning} as described in  Section \ref{sec:base_frame}. We take the AdaptaBERT as our baseline in the cross-domain scenario.
          
         \noindent \textbf{AdaptaBERT + translation}. Another baseline is set for the cross-lingual scenario. We apply translations of source data\footnote{We translate the source language data into the target language following \cite{wu2020unitrans} using MUSE \cite{conneau2017word}.} to both \textit{domain-tuning} and \textit{task-tuning} of AdaptaBERT.
         
        \noindent \textbf{TOF}. Our method is built on two baselines for the cross-lingual and cross-domain scenario, respectively. `w/o continual learning' denotes the framework without \textit{Step-6$\sim$9}. `w/o pseudo data \& w/o continual learning' denotes the framework only performs MRC enhancing at \textit{Step-1$\sim$3} of Figure \ref{fig:framework}.


\section{Results and Analysis} \label{sec:results} 

  \subsection{Overall Performance}
   
  Table \ref{tab:overall} lists main results of our method in contrast with previous state-of-the-art methods in both cross-lingual and cross-domain scenarios. 
  
   \begin{table*}[]
     \centering
    \scalebox{0.90}{
     \begin{tabular}{c|l|cccc|cccc}
     \toprule
         \multirow{2}*{\textbf{\#}}& \multirow{2}*{\textbf{Methods}} & \multicolumn{4}{c}{\textbf{cross-lingual}} & \multicolumn{4}{c}{\textbf{cross-domain}} \\
     \cline{3-6} \cline{7-10}
       ~ & ~ & es & nl & de & avg  & CBS & Twitter & WNUT16 & avg \\
     \midrule
        0 & TOF$_{MRC-enhancing}$ &  78.32 & 80.30 & 73.61 & 77.41 & 75.34 & 66.18 & 66.45 & 69.32 \\
     \midrule
         1 &  w/o target MRC data & 0.34 $\downarrow$ & - & 0.72 $\downarrow$ & - & 0.44 $\downarrow$ & 1.99 $\downarrow$ & 0.39 $\downarrow$ & 0.93 $\downarrow$ \\
         2 &  w/o source MRC data & 1.82 $\downarrow$  & 1.26 $\downarrow$ & 1.96 $\downarrow$ & 1.68 $\downarrow$ & 0 & 0.25 $\downarrow$ & 0.78 $\downarrow$ & 0.54 $\downarrow$ \\
         3 &  w/o source MRC data (trans)& 2.57 $\downarrow$ & 0.05 $\downarrow$ & 1.65 $\downarrow$  & 1.42 $\downarrow$ & - & - & - & -\\
      \midrule
         4 & w/o NER-MRC data & 3.69 $\downarrow$  & 0.83 $\downarrow$ & 1.48 $\downarrow$ & 2.00 $\downarrow$ & 0.88 $\downarrow$ & 0.65 $\downarrow$ & 0.26 $\downarrow$ & 0.59 $\downarrow$ \\
         5 & w/o NER-MRC data (trans) & 3.02 $\downarrow$ & 1.01 $\downarrow$ & 1.30 $\downarrow$ & 1.77 $\downarrow$ & - & - & - & - \\     
     \bottomrule
    \end{tabular}}
    \caption{Ablation study for TOF$_{MRC-enhancing}$, which only performs step 1$\sim$3 in Figure \ref{fig:framework}. Row 1$\sim$5 list the performance changes compared with Row 0. `$\downarrow$' denotes the drop of performance.}
     \label{tab:ablation}
  \end{table*}

  \noindent \textbf{Cross-Lingual}. 
  Our \textit{baseline} on three cross-lingual benchmarks is implemented by training AdaptaBERT with additional translations of source language data, referred as `AdaptaBERT+translations' in Table \ref{tab:overall}.
  Our method achieves significant improvements over \textit{baseline} of F1-scores 4.17, 2.49, and 4.1 for \textit{es}, \textit{nl}, and \textit{de}, respectively. 
  Compared to previous methods, our TOF framework achieves the new state-of-the-art results on two benchmarks \textit{es} and \textit{de}.
  Besides, Table \ref{tab:overall} shows the results of our TOF after removing `pseudo data'  and `continual learning', respectively, which demonstrates the effectiveness of these two enhancing strategies.
  The improvement of our TOF on \textit{nl} ($2.49 \uparrow$) is not as good as other two languages (\textit{es}:$ 4.17 \uparrow$ and \textit{de}: $4.1\uparrow$), which results from the scarcity of MRC data in \textit{nl}.
  The results well demonstrate the effectiveness of our proposed framework, which benefit from our four guidelines.

  \noindent \textbf{Cross-Domain}.
  We regard re-implemented results of AdaptaBERT as our \textit{baseline}, since it not only achieves the state-of-the-art performance on WNUT16, but also outperforms the previous state-of-the-art methods on both CBS and Twitter.
  Our framework yields obvious improvements over the \textit{baseline} (CBS: $1.11 \uparrow$, Twitter: $2.33 \uparrow$ and WNUT16: $4.83 \uparrow$) and achieves new state-of-the-art results on three datasets.
  In conclusion, all these results verify the effectiveness and generalizability of our TOF in cross-domain setting.
  

  \subsection{Ablation Study}
  We conduct ablation studies to explore how MRC datasets make difference at step 1$\sim$3 in Figure \ref{fig:framework}. Table \ref{tab:ablation} highlights the impact of different MRC data in both cross-lingual and cross-domain scenarios.
  
 \begin{table*}[]
      \centering
      \begin{tabular}{c|cccc|cccc}
      \toprule
           & \multicolumn{4}{c}{\textbf{cross-lingual}} & \multicolumn{4}{c}{\textbf{cross-domain}} \\
     \cline{2-5} \cline{6-9}
           & es & nl & de & avg & CBS & Twitter & WNUT16 & avg \\
      \midrule
       MLM $\rightarrow$ MRC $\rightarrow$ NER & 78.32 & 80.30 & 73.61 & 77.41 & 75.34 & 66.18 & 66.45 & 69.32 \\
       MRC $\rightarrow$ MLM $\rightarrow$ NER & 73.17 & 81.06 & 73.06 &  75.76 & 74.90 & 65.87 & 66.13& 68.97 \\
      \bottomrule
      \end{tabular}
      \caption{Results of our TOF framework with different fine-tuning orders.}
      \label{tab:Task_Order}
  \end{table*}
  
  In the cross-lingual scenario, we consider five kinds of MRC data: 1) `w/o target MRC data' denoting training without MRC data in the target language; 2) `w/o source MRC data' denoting training without MRC data in English; 3) `w/o source MRC data (trans)' denoting without translating the source MRC data into the target language; 4) `w/o NER-MRC data' denoting without transforming the NER data into MRC format; and 5) `w/o NER-MRC data (trans)' denoting without translating the NER-MRC data into the target language. 

  Results demonstrate that removing any data generally causes a performance drop. Therefore, we draw more in-depth observations as follows.
  For \textit{es}, `NER-MRC data' brings the greatest drop of the performance (Row 4). For \textbf{nl} and \textit{de}, `source MRC data' has the greatest impact (Row 2).  
  Besides, `source MRC data' affects the performance more than `target MRC data' (Row 2 vs. Row 1). We think it is because `source MRC data' is twice as much as the target one. 

  In the cross-domain scenario, since all of three target datasets are in English but in different domains, we do not consider the translated data (Row 3 and 5) in Table \ref{tab:ablation}. Therefore, we conduct ablation studies on three kinds of data: 1) `w/o target MRC data' denoting training without the target domain MRC data; 2) `w/o source MRC data' denoting without the source domain MRC data; and 4) `w/o NER-MRC' data denoting without transforming the NER data into MRC format. 
  
  According to the average results in Table \ref{tab:ablation}, we observe that `target MRC data', `NER-MRC data',  and `source MRC data' are in descending order of impact. 
  It is intuitive that on Twitter, as shown in Table \ref{tab:ablation} (Row 1 vs. Row 2 and Row 4), `target MRC data' has the greatest impact on the performance, when the amount of the target MRC data is greater than or equal to that of `source MRC data' and `NER-MRC data'. However, CBS is affected most by `NER-MRC data' (Row 4), since its target MRC data are collected from news domain, not science and technology news. For WNUT16, both `target MRC data' and `source MRC data' bring more drops than `NER-MRC data' (Row 1 and Row 2 vs. Row 4). We conjecture that since WNUT16 is an entity span detection task rather than standard NER, it is affected more by the golden MRC data than NER-MRC data.  
  
    \subsection{Impact of Task Order}
  We explore the impact of two different sequences for MRC-enhancing, i.e, `MLM $\rightarrow$ MRC $\rightarrow$ NER' and `MRC $\rightarrow$ MLM $\rightarrow$ NER' as shown in Table \ref{tab:Task_Order}. The results demonstrate that the former outperforms the latter in both cross-lingual and cross-domain scenarios. We conjecture that MLM can capture knowledge of data itself, e.g., domain-specific information and linguistic characters, and MRC captures task-specific information with annotations. Besides, MRC is more relevant to NER than MLM according to task relevance. Therefore, MRC is appropriate to be an intermediate task.

  \subsection{Comparison with SpanBERT}
   We replace  the pre-trained language model BERT with SpanBERT \cite{joshi-etal-2020-spanbert}  in the AdaptaBERT and MRC-enhancing of our TOF to compare the span-enhancing method with ours. The results are shown in Table \ref{tab:span_bert}.
  1) SpanBERT is superior to BERT-base for NER task (Row 3 vs. Row 1). Different from BERT with masking different tokens for each instance, SpanBERT masks a span with several adjacent tokens, which is more related to NER task.
  2) `SpanBERT' underperforms `AdaptaBERT+MRC-enhancing' on CBS and WNUT16 (Row 2 vs. Row 3), which suggests that although SpanBERT is trained on a large amount of corpus, it is not appropriate for some specific domains. Our MRC-enhancing method uses limited MRC data but achieves more improvements, which shows that MLM can not capture enough task-specific information and it is necessary to introduce other NER-related tasks.
  3) Our MRC-enhancing method can make further improvements based on SpanBERT (Row 4 vs. Row 3).
     \begin{table}[]
      \centering
      \scalebox{0.78}{
      \begin{tabular}{c|l|cccc}
      \toprule
          \#& Methods & CBS & Twitter & WNUT16 & avg \\
      \midrule
       1& AdaptaBERT & 75.30 & 65.61& 63.03 & 67.98 \\
       2 & +MRC-enhancing & 75.34 & 66.18 & 66.45 & 69.32 \\
      \midrule
      3 & SpanBERT  & 75.37 & 67.11 & 64.17 & 68.88 \\
      4 & +MRC-enhancing  & 75.48 & 67.46 & 67.80 & 70.25 \\
      \bottomrule
      \end{tabular}}
      \caption{AdaptaBERT vs SpanBERT \cite{joshi-etal-2020-spanbert} in cross-domain NER.}
      \label{tab:span_bert}
  \end{table}
  
   \section{Related Work}
  \noindent \textbf{Zero-resource NER}. 
  Some studies \cite{jia-zhang-2020-multi,pfeiffer-etal-2020-mad, vidoni2020orthogonal} focus on improving architectures of existing models, which add new components into networks to capture specific knowledge, i.e., entity types, language and task characteristics. 
  Different from these methods, our approach only modifies the training procedure without changing model structures. 
  Other studies introduce different auxiliary tasks to alleviate data scarcity \cite{han-eisenstein-2019-unsupervised,  xue2020coarse, phang2020english}. They are usually based on multi-task learning or two-phrase fine-tuning. Multi-task learning requires balance between the target task and auxiliary tasks, which needs carefully designed objectives. Although two-phrase fine-tuning is effective, it is still inadequate for available data and depends on valid data selection. 
  Our work differs in that we not only propose four practical guidelines to guide data selection and task fine-tuning, but also design a task-oriented fine-tuning framework to exploit more diverse data and target-oriented training strategies. 
  

  \noindent \textbf{Data Augmentation}. Our approach is inspired by some studies on text classification. \citet{gururangan-etal-2020-dont} utilize unlabeled data in different domains and tasks. \citet{ben-david-etal-2020-perl} exploit unlabeled corpora from multiple domains. Unlike these methods, we focus on target domain/language data with annotations for other tasks, which not only transfers domain/language knowledge, but also utilizes available annotations for other tasks.
  
  \noindent \textbf{MRC for Different Tasks}. Although most researches on NER focus on the sequence labeling framework \cite{huang2015bidirectional, ma2016end, akbik2018contextual, liu-etal-2019-gcdt}, our work is also inspired by formatting other tasks as MRC, such as NER \cite{li-etal-2020-unified}, co-reference resolution \cite{wu2019coreference}, and event extraction \cite{liu2020event}. These studies show the superiority and scalability of MRC framework and provide a reference for our work. Different from \cite{li-etal-2020-unified} using MRC to build a new solution architecture for NER, we exploit MRC to improve the training procedure of NER that is based on sequence labeling. Besides, we perform continual learning between MRC and NER to enhance the impact of MRC on NER.

\section{Conclusion and Future Work}
In this paper, we analyze the problem of data scarcity in zero-resource NER. To alleviate this issue, we propose four practical guidelines on transferring knowledge from three aspects, i.e., domain, language, and task, and strengthening connections between the source and target data. Based on these guidelines, we design a task-oriented fine-tuning framework to enhance the training procedure with various strategies.
Our approach yields significant improvements on six benchmarks and achieves the state-of-the-art on five benchmarks.
In the future, we will extend our framework on different target tasks and more task-specific enhancing strategies.

\section*{Acknowledgments}
The research work descried in this paper has been supported by the National Nature Science Foundation of China (No. 61976016, 61976015, and 61876198) and the National Key R\&D Program of China (2020AAA0108001). The authors also would like to thank the anonymous reviewers for their valuable comments and suggestions to improve this paper.

\bibliographystyle{acl_natbib}
\bibliography{acl2021}

\appendix

\section{Statistics} \label{sec:statistics}
The statistics of all datasets are listed in Table \ref{tab:datasets}.  
We regard CoNLL03 in English as the source NER data in both cross-lingual and cross-domain scenarios.
For target NER datasets, we consider cross-lingual and cross-domain scenarios, respectively.

In the cross-lingual scenario, CoNLL03 in German, CoNLL02 in Spanish, and CoNLL02 in Dutch denote the benchmark datasets in the target languages, i.e., German (\textit{de}), Spanish (\textit{es}), and Dutch (\textit{nl}),  respectively. In terms of MRC datasets, we apply MLQA in Spanish and XQuAD in German as labeled MRC datasets in the target languages, i.e., on \textit{es} and \textit{de}.  Note that we use the initial validation and test splits in MLQA and XQuAD as the training and validation sets in our work. Since it is difficult to obtain labeled MRC datasets for \textit{nl},  we consider the MRC data in the source language, i.e., English (\textit{en}).

In the cross-domain scenario, CBS SciTech News NER datasets, short as CBS, in science and technology news domain, Tiwtter NER dataset in twitter domainm, and the shared task on entity span detection for WNUT2016 in twitter domain are considered as the target domain NER datasets. All of these three cross-domain benchmarks are in English. We use NewsQA in news domains for MRC fine-tuning on CBS, due to lack of available MRC data in science and technology news domain. TweetQA is applied to both Twitter and WNUT16 NER as the MRC data in the target domain.

\begin{table*}[]
         \centering
         \scalebox{0.95}{
         \begin{tabular}{lcccrrr} 
            \toprule
            \textbf{Benchmark\quad}  &  \textbf{\quad Task\quad} & \textbf{Language} & \textbf{Domain} & \textbf{Training} &  \textbf{Validation} & \textbf{Test}\\
            \midrule
              CoNLL03  & NER & English(en) & News & 14987 & 3466 & 3684 \\
            \midrule
             \multicolumn{7}{c}{\textbf{Cross-Lingual}} \\
            \midrule
             
             CoNLL03  & NER & German(en) & News & 12705 & 3068 & 3160 \\
             CoNLL02  & NER & Spanish(es) & News & 8323 & 1915 & 1517 \\
             CoNLL02  & NER & Dutch(nl) & News & 15806 & 2895 & 5195 \\
             MLQA  & MRC & Spanish(es) & Multi-domain & 5254 & 500 & - \\
             XQuAD  & MRC & German(de) & Multi-domain & 1190 & 428 & - \\
             SQuAD & MRC & English(en) & Multi-domain & 10000 & 1000 & - \\
             \midrule
              \multicolumn{7}{c}{\textbf{Cross-Domain}} \\
             \midrule
             CBS SciTech  & NER & English(en) & Science-technoloy news & - & - & 2000 \\
             TwitterNER  & NER & English(en) & Social-media &  4000 & 1000 & 3256 \\
             WNUT16  & ESD$^{*}$ & English(en) & Social-media &  2394 & 1000 & 3856 \\
             NewsQA  & MRC & English(en) & News &  92550 &  5167 & 5127 \\
             TweetQA  & MRC & English(en) & Social-media &  10692 & 1086 & 1979 \\
            \bottomrule
         \end{tabular} }
         \caption{Dataset statistics. `$^{*}$' denotes the entity span detection task.  }
         \label{tab:datasets}
 \end{table*}
     
 \section{Hyperparameters} \label{sec:hyperparameter}
  \begin{table*}[]
    \centering
    \begin{tabular}{c|c|ccc|ccc} 
        \toprule
          \multirow{2}*{\textbf{Methods}}& & \multicolumn{3}{c}{\textbf{cross-lingual}} & \multicolumn{3}{c}{\textbf{cross-domain}} \\
         \cline{4-5} \cline{6-8}
         ~ & & es & nl & de & CBS & Twitter & WNUT16 \\
         \midrule
         MRC-enhancing & MRC & 2e-6 & 1e-6 & 1e-6 & 2e-5 & 2e-5 & 2e-5 \\
         MRC-enhancing & NER & 5e-5 & 2e-5 & 2e-5 & 5e-5 & 2e-5 & 5e-5\\
         Pseudo data & NER & 2e-5 & 1e-5 & 1e-5 & 8e-5 & 1e-6 & 1e-6\\
         Continual learning & MRC & 5e-5 & 2e-6 & 2e-5 & 8e-6 & 3e-5 & 1e-6\\
         Continual learning & NER & 5e-5 & 5e-6 & 1e-6 & 3e-5 & 5e-6 & 2e-6\\
         \bottomrule
    \end{tabular}
    \caption{Learning rate of MRC and NER model on different datasets.}
    \label{tab:hyper}
\end{table*}
 
In the training procedure, we fine-tune each MLM model for 3 epochs, MRC for 6 epochs, and NER for 6 epochs. Besides, we perform continual learning for one iteration and achieve the best performance. 
To re-implement our results easily, we set the seed to a fixed value as 2019 following \cite{han-eisenstein-2019-unsupervised}. The learning rate of MLM on each dataset is set to 5e-5. The learning rate of MRC and NER model on different datasets is listed in Table \ref{tab:hyper}. To find the proper learning rate, we perform the hyperparameter search on a set of learning rete values, i.e., {1e-6, 2e-6, 5e-6, 8e-6, 1e-5, 2e-5, 3e-5. 5e-5, 8e-5}. And we choose hyperparameter values according to the best validation performance. We train our model on one NVIDIA Tesla P40 (24GB). The average runtime of our TOF framework on different datasets varies from 6 hours to 2 days, due to the data size of different datasets. Other hyperparameters are set following \cite{han-eisenstein-2019-unsupervised} and \cite{li-etal-2020-unified}.

\end{document}